# Neighborhood Representative for Improving Outlier Detectors

Jiawei Yang*, Yu Chen*, Sylwan Rahardja *

*Abstract* **—Over the decades, traditional outlier detectors have ignored the group-level factor when calculating outlier scores for objects in data by evaluating only the object-level factor, failing to capture the collective outliers. To mitigate this issue, we present a framework called *neighborhood representative* (NR), which empowers all the existing outlier detectors to efficiently detect outliers, including collective outliers, while maintaining their computational integrity. It achieves this by selecting representative objects, scoring these objects, then applies the score of the representative objects to its collective objects. Without altering existing detectors, NR is compatible with existing detectors, while improving performance on eleven real world datasets with +8% (0.72 to 0.78 AUC) on average relative to twelve state-of-the-art outlier detectors. The implementation of NR can be found via www.OutlierNet.com for reproducibility.**

## I. Introduction

*Outliers* are objects in a dataset that significantly deviate from other objects [1]. Detecting anomalous data in large datasets has many potential applications [1] such as, abnormal time-series [2], traffic patterns [3, 4], and fraud detection [5]. Outliers can be considered as *errors*, affecting statistical algorithms like clustering [6]. Recently, outlier detection has also been applied to wireless sensor networks [7], attributed networks [8], information networks [9], social networks [10], point cloud registration [11], high-dimensional settings [12], and video data [13]. These highlight the importance of outlier detection and its diverse applications at present.

Most outlier detectors calculate the *outlier score* for every object independently, and then objects with scores exceeding a threshold [14] will be classified as outlier objects. To improve the results of multiple outlier detectors, *ensemble techniques* [15,16,17] have been developed to combine multiple detectors to obtain more accurate results [1]. To improve the results of a single outlier detector [18, 19], Yang et al. hypothesized that similar objects should have similar outlier scores and proposed a post-processing technique called neighborhood averaging to improve individual outlier detector.

*equally contribute to the paper
Jiawei Yang is with the School of Marine Science and Technology, Northwestern Polytechnical University, 127 West Youyi Road, Xi'an, Shaanxi 710072, China. (jiaweiyang@ieee.org).
Yu Chen is with Neuroscience Institute & Machine Learning Department, Carnegie Mellon University, Pittsburgh PA, USA. (yuc2@andrew.cmu.edu)
Sylwan Rahardja is with University of Eastern Finland, Joensuu, Finland (sylwanrahardja@ieee.org)

A typical process of detecting outliers is summarized in Fig. 1. For a typical outlier detector, the features of input X can be pre-processed, such as by scaling or normalization, before the outlier detector analyses the data. Next, detectors calculate the *outlier score* for each object, which is a score used to determine the probability of an object being an outlier. Usually, the outlier score is post-processed to achieve robust results, such as by ensemble techniques or neighborhood averaging [18, 19]. Finally, objects with large scores are classified as outliers [1]. The accuracy of the outlier score is critical to the success of outlier detection.

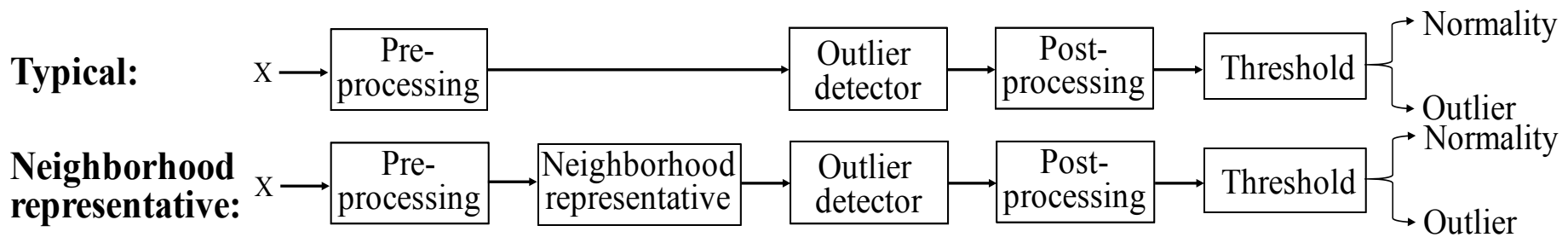

Fig. 1. Outlier detection process with and without Neighborhood Representative.

Despite rigorous research into the field of outlier detection, the traditional outlier detectors still suffer from a major limitation. When computing outlier score for an object in a dataset, outlier detectors only consider the individual objects, but fail to analyze a group of objects in the region surrounding a particular object. Hence, they only evaluate the factor of object-level without considering the factor of group-level by which the objects are near or belong to. This limitation is exacerbated in datasets with many collective outliers as demonstrated by objects G2 in Fig 2, severely impairing the performance of traditional outliers.

Original data

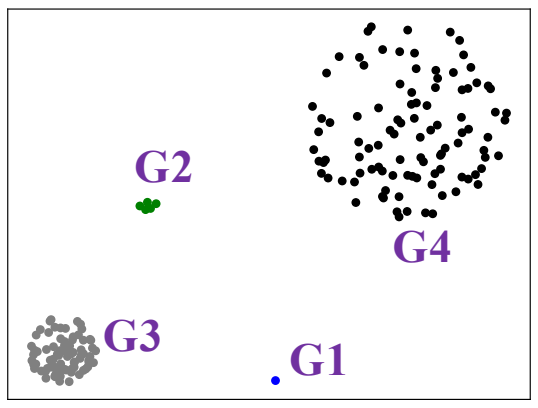

(1)

A case of three detectors

| Groups | G1 | G2 | G3 | G4 |
|---|---|---|---|---|
| Objects | 1 | 5 | 80 | 100 |
| Detector 1 | 1 | 1 | 1 | 1 |
| Detector 2 | 19 | 12 | 1 | 1 |
| Detector 3 | 27 | 21 | 10 | 2 |

Mean outlier scores for different groups of objects

(2)

A case of a detector with and without NR or NR+

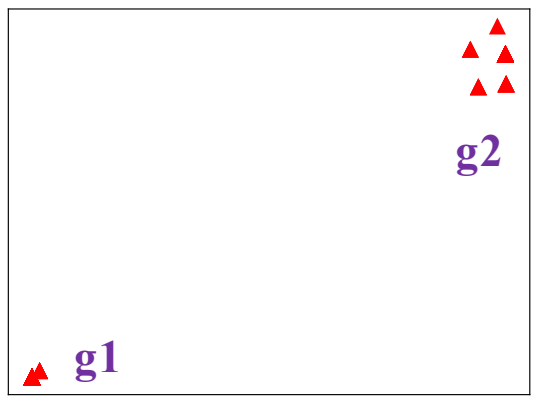

(3)

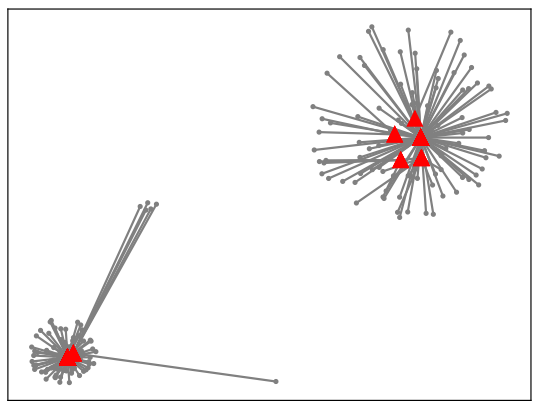

(4)

| Groups | G1 | G2 | G3 | G4 |
|---|---|---|---|---|
| Original | 19.06 ±0.00 | 12.73 ±0.00 | 1.04 ±0.07 | 1.03 ±0.07 |
| NR | 8.52 ±0.03 | 8.52 ±0.03 | 8.52 ±0.03 | 0.99 ±0.06 |
| NR+ | 27.58 ±0.03 | 21.25 ±0.03 | 9.56 ±0.09 | 2.02 ±0.13 |

Mean ± standard deviation

(5)

Fig. 2. (1): original data with four groups of objects: G1 (global outlier), G2 (collective outliers), G3 (normalities), and G4 (normalities). (2). An analogy of mean outlier scores for different groups of objects from three detectors. (3): Selected representatives (medoid object) for the original data in (1). (4): Representation of each object by adjacent medoid object. (5): Mean outlier scores of objects in these four groups from detector LOF [20] without (labeled as *original*) and using NR or NR+.

The dilemma posed by object clustering in datasets can be illustrated by a hypothetical scenario in Fig. 2. (2). In Fig. 2. (1), there are four groups of objects G1, G2, G3, and G4 with 1, 5, 80 and 100 objects respectively. The single object in G1 is called a *global outlier* while the objects in G2 are called *collective outliers* [1]. As shown in Fig. 2. (2), detector 1 could not discriminate all data points and was thus ineffective. In contrast, detectors 2 and 3 were able to classify G1 and G2 with higher scores and thus identify G1 and G2 as outliers. On the other hand, the classification of G3 and G4 as outliers was controversial as they both had clusters of data objects, hence G3 and G4 could be relative outliers depending on the rest of the dataset. Detector 3 was favored over 1 and 2 due to its ability to identify G3 and G4 as possible outliers. The ability to accurately classify outliers, regardless of size and region, is key to improvement of outlier detectors. This will be further substantiated in the following sections.

We further analyzed another case shown in Fig. 3 where 2 groups of objects had rather similar mean outlier scores like G3 and G4 in Fig. 2. (5). Similar to the case in Fig. 2, both normal groups and outlier groups had a large collection of data points. Outlier detectors, such as local object factor (LOF) [20] and one class support vector machines (OCSVM) [21], erroneously classified the objects in collective outlier group as normalities due to the large concentration of the data. A clear example in Fig. 3 was illustrated by object q. It was harder to detect object q as the outlier as compared to object p, as it could be visually established that while object p was a large distance from the true normalities, q was nested within a group of true outliers.

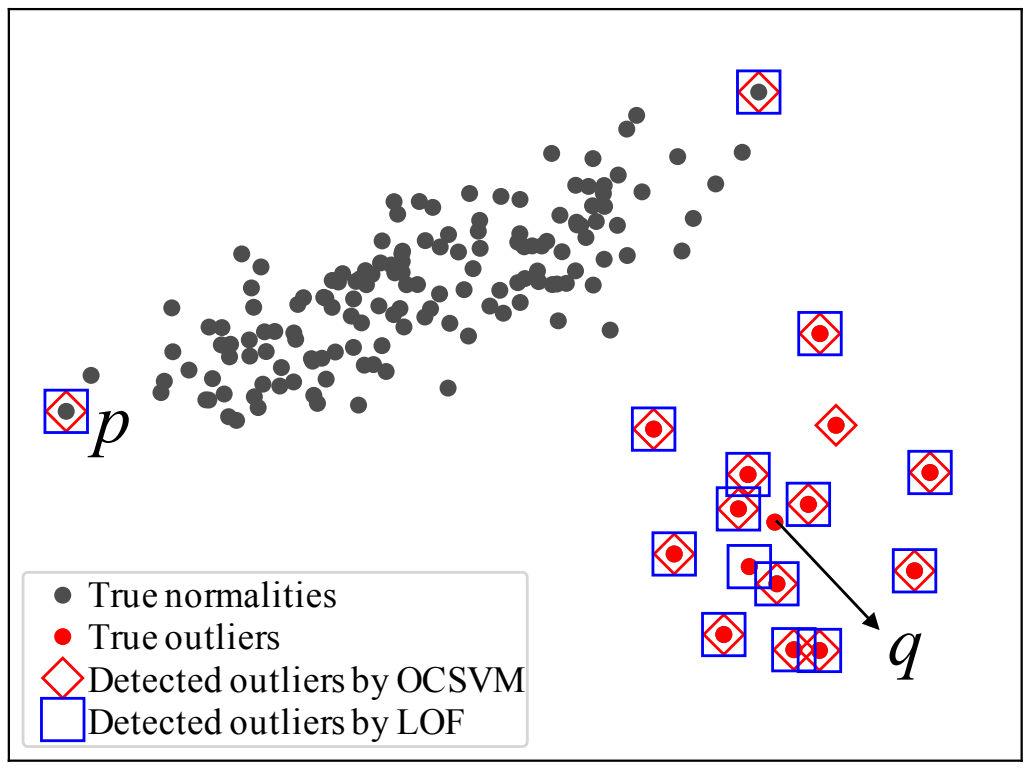

Fig. 3. An example of outlier detection. The output of LOF was labeled by squares. The output of OCSVM was labeled by diamonds. Both detectors had inadequacies. Object *p* was a false negative. Object *q* was a false positive.

In light of the limitations of prior models, we propose a framework called *Neighborhood Representative* (NR) and its extended version NR+. NR aims to provide a preprocessing tool to improve any single outlier detector's ability to distinguish collective outliers. The proposed outlier framework considers both individual position of datasets and adjacent objects. NR works in three steps: First, a representative object for several collective objects is selected based on *k*-nearest neighbors (*k*-NN). Secondly, the representative object is scored. Thirdly, the score of the representative object is taken as the score of its collective objects.

NR works in sequential steps as shown at the bottom of Fig. 1. In contrast to typical outlier detection methods, using NR has the additional step of data preprocessing prior to usage of outlier detectors. The key difference between the proposed NR and existing detectors is that NR is a framework for improving any exiting single outlier detector, thus capitalizing on existing detectors to improve their respective performance.

We postulate that the proposed framework NR and NR+ will revolutionize outlier detection in four key aspects. First, this is a new and potentially revolutionary assumption for outlier detection. Secondly, we utilize the group-level factor when considering when classifying outliers. Thirdly and most significantly, NR and NR+ aims to improve all score-based outlier detectors. Fourthly, NR is the first preprocessing method which capitalizes on existing detectors to improve the performance of all existing detectors.

## II. Concepts related to outliers and outlier detectors

This section aims to do a detailed review of existing state-of-the-art detectors, their assumptions and limitations.

### A. *Concepts related to outliers*

Clustering and outlier detection are two closely related data mining tasks. In some literature, outlier detection is seen as a vice product of clustering. However, this is not true. There are two main reasons for this assertion [22]: Firstly, clustering algorithms are optimized to find clusters rather than outliers. Secondly, a set of many outlier objects organized within close proximity as seen in Fig. 2 and Fig. 3 will be recognized as clusters rather than collective outliers. For example, a clustering algorithms will erroneously classify the true normalities and true outliers in Fig. 3 as two separate clusters, thus having a subtle but important difference in output as compared to outlier detectors.

Outlier detection also has many assumptions and definitions. Hawkins [23] defines *an outlier as an observation that deviates so much from the other observations to arouse suspicions generated by a different mechanism*. Barnet et al. [24] stated that *an outlying observation, or outlier, is one that appears to deviate markedly from other members of the sample in which it occurs*. Johnson [25] mentioned that an outlier is *an observation in a data set which appears to be inconsistent with the remainder of that set of data*. Some other assumptions are either based on the data structure or the outlier detectors used. These definitions vary slightly, but have the common limitation of ignoring collective outliers or group-level outliers. This limitation will be evident when applied to the problem of detecting collective outliers in Fig. 2. (2) and Fig. 3. In contrast, the proposed algorithm does not disregard presence of possible groups of outliers, with the goal of accurately identifying any existing collective outliers.

### B. *Outlier detectors*

Extensive research had been done for outlier detectors, with a plethora of outlier detectors proposed. While the principles of outlier detection are conserved, each proposed detector had its unique aspect and advantage.

Distance-based outlier detectors rely on the distance between data objects, where outliers are datasets with no adjacent or surrounding objects. Ramaswamy et al. [26] calculated the distance between an object and its $k^{th}$ neighbor to give an outlier score. We refer to this detector as *k*-NN. Its variant calculates the average distance to its all *k* neighbors [27]. Instead of using *k*-NN, Knorr et al. [28] defined a distance threshold and calculated the number of objects within the given distance to the object, using this score as the outlier score.

Hautamäki et al. [27] proposed a method called *outlier detection using indegree of nodes* (ODIN) using the *k*-NN graph, but rather than calculating the distances between objects, they calculated the in-degree as the outlier score.

Li et al. [29] developed a detector called *reverse unreachability*, which was a representation-based detector. A given object was represented by a linear combination of its *k*-nearest neighbors. Their proposed representation provided a weight matrix of the relative contribution of each neighbor. Negative weight referred to an abnormal relation between neighbors. The outlier score was the number of negative weights.

Density-based detectors postulated that the density of the outliers were considerably lower than the density of their neighbors. *Local outlier factor* (LOF) [20] calculated the relative density of an object to its *k*-nearest neighbors and used this as an outlier score. LOF emerged as the best detector amongst twelve *k*-NN detectors [36,30].

The statistical-based detectors assumed that normality objects follow the same distribution, whereas outliers do not. The *minimum covariance determinant* (MCD) [31] looked for 50% of objects with the smallest scatter. The outlier score was the distance between an object and the center of the defined 50% of objects.

Tree-based detectors fundamentally differed from the other detectors that rely on distance or density. As defined by Liu et al. [32], *isolation-based anomaly detection* (IFOREST) constructed a tree by partitioning the dataset. It recursively split the data by a randomly selected split value between the maximum and minimum of a randomly selected feature. The average of the path lengths over the random trees was the outlier score. Recently, an improved version of IFOREST was developed by Tan et al. [33].

Support vector machines recognized patterns in data and can be used in classification tasks. *One class support vector machine* (OCSVM) [21] trained the support vector model by treating all objects as one-class. The outlier score was the distance between an object to the model.

Some detectors modified the data and measured the distance between an object and its modified version as the outlier score. Different outlier detectors have different modification techniques.

*Mean-shift outlier detection* (MOD) [34, 35] modified the data by replacing an object with the mean of its *k*-NN. This process was repeated three times. The distance between the original and the modified value of an object was the outlier score. Its extended version MOD+ [36] used the sum of shift distances of *k*-NN as outlier score. This approach was most effective when the amount of outliers was large [34].

*Principal component analysis* was an established technique in data mining. It revealed the data structure and explained the variance. It worked by extracting the principal features of datasets. The *principal-component-analysis-based outlier detection method* (PCAD) [37] modified the data by computing the projection of an object on the eigenvectors with a normalized reconstruction error. The error was interpreted as the outlier score.

Variational AutoEncoder (VAE) aimed to provide an automated discovery of interpretable factorized latent representations. It used reconstruction error as the outlier score [38]. Beta-VAE [39] was a type of VAE and uses the combination of reconstruction error and the non-negative Kullback–Leibler divergence between the true and the approximate posterior as the outlier score.

Single-objective generative adversarial active learning (SO-GAAL) [40] was a neural network trained to classify its generative data and real data. The possibilities assigned to objects in real data were interpreted as as outlier scores.

Copula-based outlier detector (COPOD defined in [41]) was a parameter-free outlier detector to detect outliers in tails. It constructed an empirical copula to predict tail probabilities of each object in each single dimension and used the probability as outlier score. Not limited to outliers in tails, Yang et al. proposed a method to detect outliers both in tails and between clusters [42].

The detectors mentioned above are based on different assumptions but have a common limitation of computing outlier score for objects individually.

## III. METHODOLOGY

In this section, we introduce an assumption for outliers and describe the propose methods NR and NR+. We also discuss their conceptually related techniques in literature.

### A. Proposed assumption for outliers

Given a data set containing no duplicates and comprising of several group of objects, we assumed that *the groups of objects with fewer objects were more likely to be outliers*. Given a data set X, we assumed that:

$$p(x) > p(y) \text{ if } x \in G^L,\ y \in G^E \text{ and } |G^L| < |G^E| \quad (1)$$

where $G^*$ is a group of objects and X=$G^1 \cup G^2 \cup \ldots \cup G^m$; p(*) is the possibility of the object being an outlier; |*| is the size of the group. This was consistent with the concept proposed by Ester et al. [43] that objects in a group having insufficient objects to form a cluster were noise or outliers.

The assumption stated in equation (1) corresponded with the outlier scores in Fig. 2 (2), because G4 had more objects than G3. Most outlier detectors did not reflect any information of the size of the group of a data object, which was discovered to be important in labeling collective outliers.

### B. Group-level factor

We approximate the group size using the relationship between the mean outlier scores between groups. With the group size information, it was relatively easier to pick out collective outliers. The relationship between mean outlier scores and the group sizes could be expressed based on empirical observations by the following equations:

$$m^D (G^L) >> m^D (G^E) \text{ if } |G^L| << |G^E| \quad (2)$$

$$m^D (G^L) \cong m^D (G^E) \text{ if } |G^L| \cong |G^E| \quad (3)$$

where $G^*$ is a group of objects and X=$G^1 \cup G^2 \cup \ldots \cup G^m$; $m^D(*)$ is mean outlier scores from detector D; |*| is the size of the group. Similarly, equations (2) and (3) could be illustrated by Fig. 2. (5). The sizes of G1 and G2 were much smaller than G3 and G4, and the mean scores of G1 and G2 were much larger. In contrast, the size and mean outlier scores of G3 and G4 do not differ significantly. The rational of this relationship was further explained in section D.

*C. Proposed framework*

We propose to preprocess the data by assigning the group-level information. Firstly, we sampled objects from each group as follows:

$$|R(G^L)| << |R(G^E)| \text{ if } |G^L| < |G^E| \quad (4)$$

where R(*) refers to *representative objects* containing the objects sampled from group $G^*$, namely R($G^*$) ⊂ $G^*$, and |*| is the size of the group.

Then, the preprocessed data could be fed to any existing outlier detectors introduced in Section II to calculate the outlier scores for object in R($G^*$); 3. The object in $G^*$, but not in R($G^*$), would share the same score of the nearby object in R($G^*$). We referred to this strategy as the unbalanced reduction strategy (URS). As a result, it could be represented by the following equation:

$$m^{URS+D}(G^L) \ll m^{URS+D}(G^E) \text{ if } |G^L| < |G^E| \quad (5)$$

where $m^{URS+D}(*)$ is the mean outlier scores calculated by detector D with the proposed URS.

The goal was to amplify the group size differences. The table in Fig. 2. (5) shows the amplification after application of the proposed framework. Such a procedure can be repeated several times.

*D. Neighborhood representative (NR)*

We found that sampling using the *medoid* object of a set of objects could be reliable. Given a set of objects X= $\{x_1, x_2 \ldots, x_k\}$, the *medoid* object is defined as the objects in X which had the minimum summed distance from other objects in X. It was defined as follows:

$$x_{medoid} = \arg\min_{y \in X} \sum_{i-1}^{k} d(y, x_i) \quad (6)$$

where d(*) is the distance function such as Euclidean distance. Fig. 4 illustrates the process of sampling medoid from a set of neighborhood objects. Given an object A, the medoid of its *k* nearest neighborhood (*k*-NN, *k*=3) objects (B, C, D) was object C. The neighborhood representative (NR) was the URS with sampling medoid from *k*-NN.

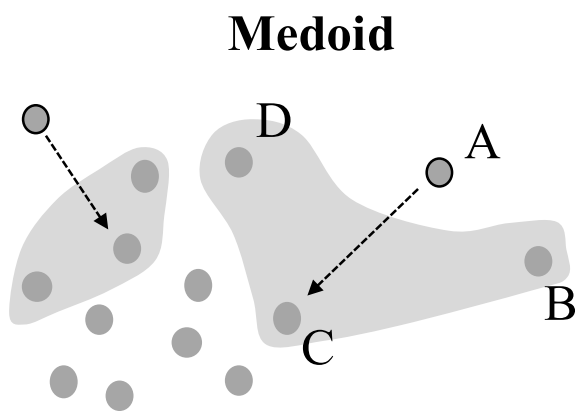

Fig. 4. medoids: object C is the medoid over objects B, C, and D.

NR employed the medoid-shift [20] technique to find the representative objects. Medoid-shift replaced each object by the medoid of its *k*-NN objects as written in Algorithm 1. Many objects within the dataset might share a common object as their medoids, making the distribution of collective outliers sparse. Upon analysis of the distribution, outlier detection was significantly more convenient. This concept could be illustrated by Fig. 5 with the same dataset used in Fig. 3 for simplicity of comparison. The blue dots were the selected representatives. The red circles indicated the outlier score values obtained from LOF [20]. A larger circle referred to a larger value and thus higher likelihood of an object being classified as outlier. Each data object (dark dot) was linked to an adjacent representative, and thus its outlier score was set as that of the representative.

**Algorithm 1:** Medoid-shift process (MS (X, *k*, I))

**Input**: Dataset X, neighborhood size *k*, iteration I
**Output**: Modified X*

REPEAT I TIMES:
FOR EACH OBJECT $X_i \in X$:
k-NN($X_i$) ← Find its *k*-nearest neighbors $\in X$;
$M_i$ ← Calculate the medoid of the neighbors *k*-NN($X_i$);
$X^*_i$ ← $M_i$ (replace $X_i$ by the medoid $M_i$ ), $X^*_i \in X^*$;

**Algorithm 2:** Neighborhood representative NR(X, *k,* D)

**Input:** Dataset X, Outlier detector D, Neighborhood size *k*
**Output:** Outlier scores S

R ← Find representative objects of X by applying MS (X, *k*, 1);
O ← Apply D to R to calculate outlier scores for representative objects;
$S_i$ ← $O_j$. $R_j$ is the representative of $X_i$, $X_i \in X$;

The pseudo-code of the proposed NR was shown in Algorithm 2. Given a dataset X and any outlier detector D, the proposed NR worked in three steps: First, for every object $X_i$, the representative object $R_j$ was found by applying medoid-shift (Algorithm 1) with one iteration as default. Secondly, we applied detector D to calculate the outlier score $O_j$ for the representative object $R_j$. Thirdly, the outlier score $O_j$ was used as the outlier score of object $X_i$. An example of using NR was shown in Fig. 6 with the same dataset used in Fig. 3 and Fig. 5. After using NR, both LOF and OCSVM could successfully detect all outliers, as a comparison to the results without using NR in Fig. 3 showing the glaring incompetency without NR. The idea of medoid-shift had been used for clustering [17, 18, 43], for removing noise from clustering [35, 36], and for creating outlier detectors [41]. In comparison, NR used medoid-shift as a pre-processing technique to select representative objects prior to outlier detection.

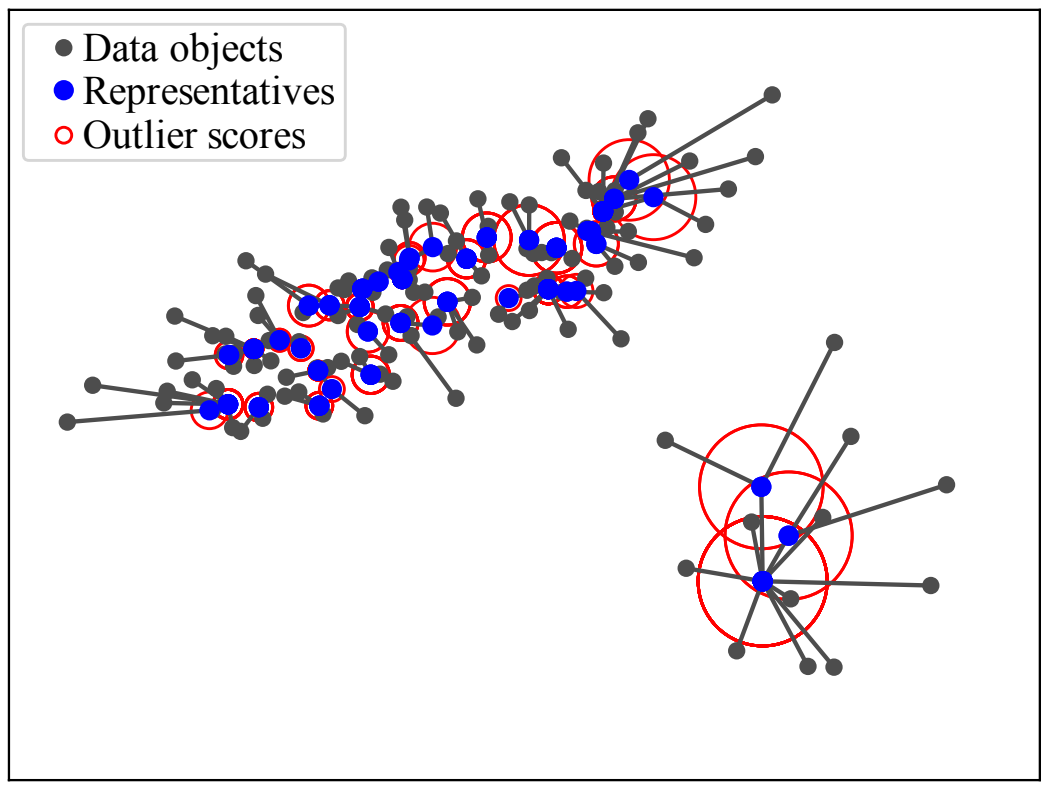

Fig. 5. Application of NR to dataset used in Fig. 3. Data was pre-processed by NR to produce a set of representatives (blue dots). Detector LOF was employed to score the representatives. The radius of the red circle indicated the value of the score. Each data object (black dots) was assigned to a representative (linked by the solid lines) and accorded similar outlier score as the representative object.

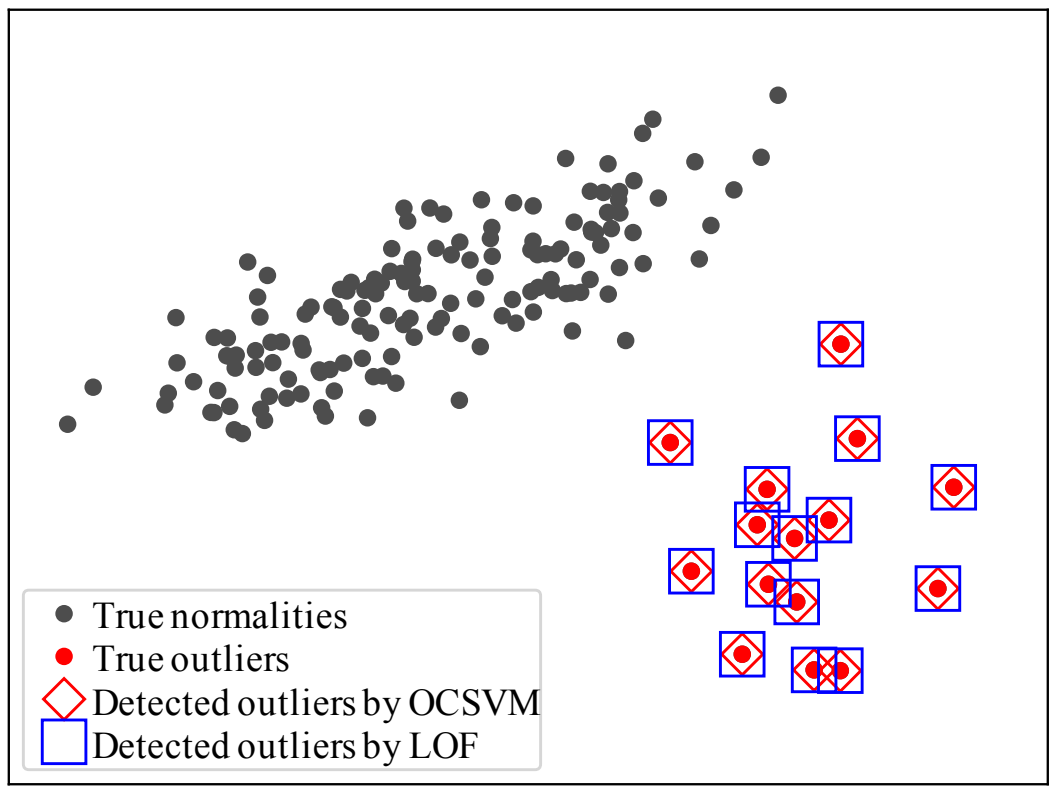

Fig. 6. Results of outlier detection by LOF detector (square) and OCSVM (diamond) after applying NR as pre-processing. The example dataset was the same as Fig. 2. (1). All the outliers were correctly detected in both cases.

To prove that the sampling medoid from *k*-NN as R(*) could reach the goal shown in Eq. (4), a rigorous theoretical proof is challenging because the medoids may be affected by many factors such as data distribution, cluster shape, cluster size, and the locality decided by the parameter *k*. To simplify the analysis, we considered two groups of objects G1 and G2 with n and m objects respectively as shown in Fig. 7. To simplify the analysis, we considered two groups of objects G1 and G2 with n and m objects respectively as shown in Fig. 7. Given an object $x_j \subset G^*$ where $G^*$means G1 or G2, we defined its representative object as follows:

$$x_{medoid} = \arg\min_{y \in X} \sum_{i-1}^{k} d(y, x_i),\ X = k\text{-NN}(x_j) \cup x_j,\ X \subset G^* \qquad (7)$$

Fig. 7 illustrated both one-dimensional and two-dimensional data. In summary, when the objects in G1 and G2 were highly symmetrical in shape, such as square-shaped in a two-dimensional space and cube-shaped in a three-dimensional space, the sampled objects, namely medoid objects in blue dot, were also highly symmetrical in shape.

To study the ratio between the number of sampling objects from G2 and G1, the number of sampling objects from G1was preset to be one by setting k for k-NN to n, which the size of G1. However, this setting did not necessarily imply that the number of sampling objects from G2 relied on the size of G1, but the decision was made to avoid a convoluted analysis. The equation could then sample $(m^{\frac{1}{d}} - n^{\frac{1}{d}} + 1)^d$, where d is the dimension of the feature space, from G2. The original disparity in size between G2 and G1 was |G2|/|G1||=m/n, and this disparity of sizes were increased to R(G2)|/|R(G1)|= $(m^{\frac{1}{d}} - n^{\frac{1}{d}} + 1)^d/1$ after sampling

medoids as representative objects. From Fig. 5, we could see that in one-dimensional space, |G2|/|G1||=2.3 but |R(G2)|/|R(G1)|=5; in one-two space, |G2|/|G1||=5.4 but |R(G2)|/|R(G1)|=25. Both cases reached the goals established by Eq. (4).

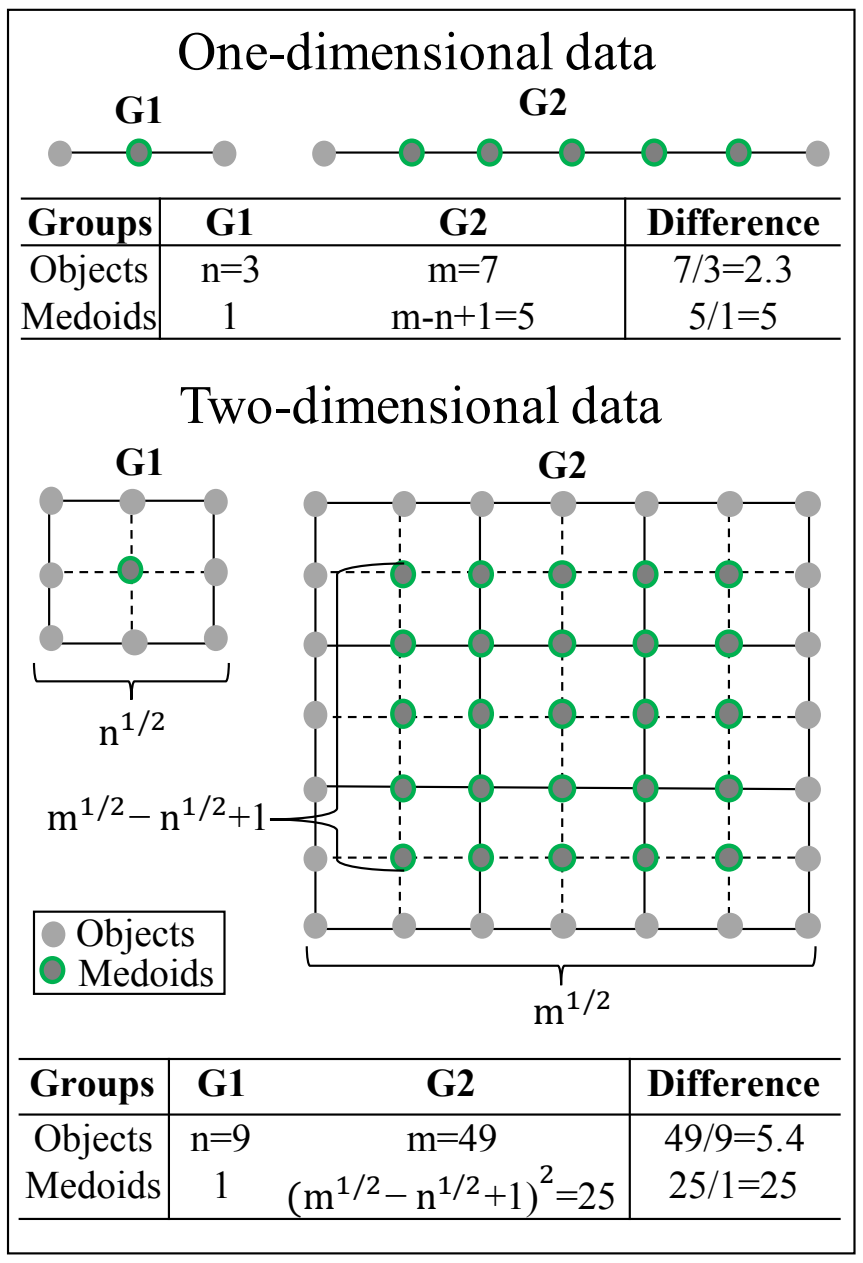

Fig. 7. An ideal case of evaluating the difference of the number of medoids between two groups of objects in one- and two-dimensional space.

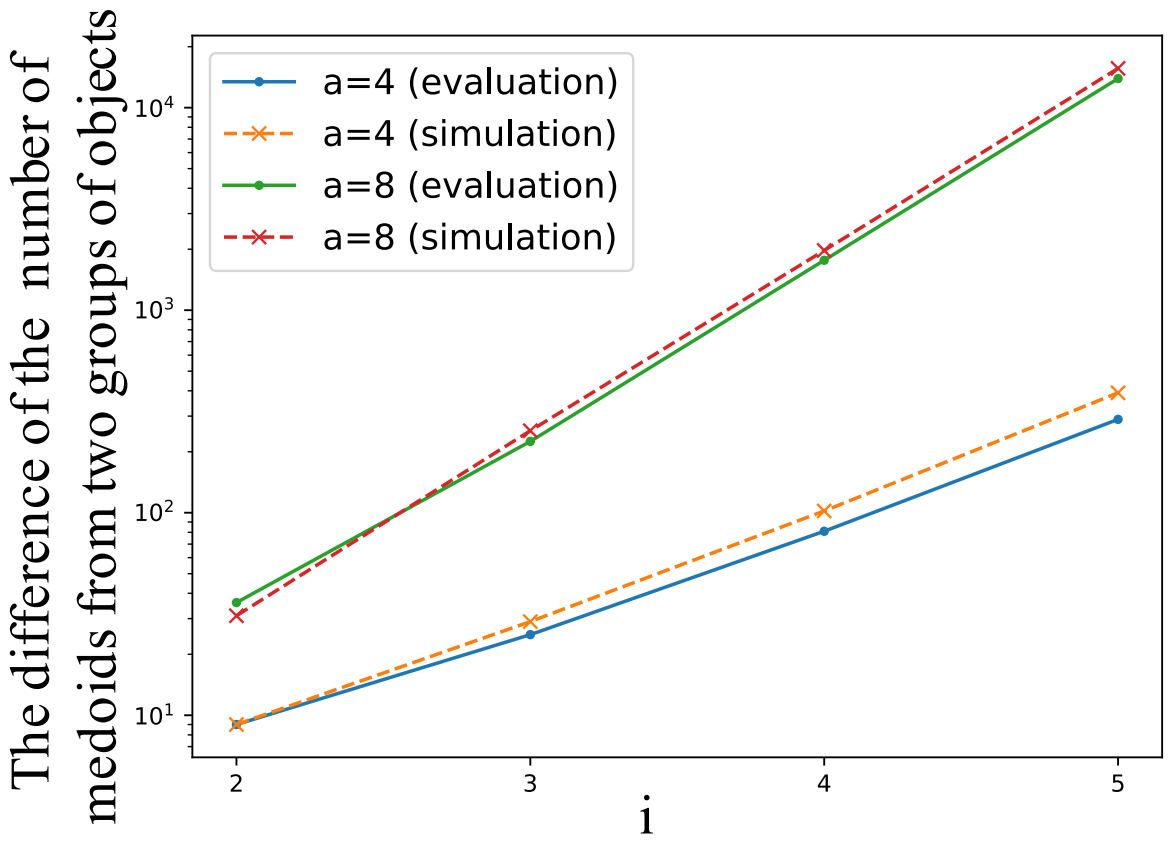

Fig. 8. The difference of the number of medoids from the groups with m=$a^i$ and n= $a^{i-1}$ objects, where i=2,3,4,5. The dash line was the exact number of medoids (*k*=n for *k*-NN) found from generated data. The solid line was the evaluated number of medoids calculated by $(m^{\frac{1}{d}} - n^{\frac{1}{d}} + 1)^d$ where d=2 for two-dimensional space.

To further understand the relationship between m/n and $(m^{\frac{1}{d}} - n^{\frac{1}{d}} + 1)^d/1$ and to facilitate plotting the relationship in a log-scale, we simulated an experiment with setting m=$a^i$, n=$a^{i-1}$ with i ranging from 2 to 5, and *k* = n. The results of the experiment were shown in Fig. 8. The dashed line shows the results of representative objects, namely medoids, via randomly generated datasets containing two groups of objects, thus reflecting the number of medoids from the data of the simulation. The solid line showed the results calculated by $(m^{\frac{1}{d}} - n^{\frac{1}{d}} + 1)^d$, thus reflecting the calculated value for reference. From Fig. 8, we deduce that the simulated results matched the reference calculated value. This deduction highlighted the accuracy of $(m^{\frac{1}{d}} - n^{\frac{1}{d}} + 1)^d$. In addition, as the value of a increased, both m and n increase and the value of |R(G2)|/|R(G1)|= $(m^{\frac{1}{d}} - n^{\frac{1}{d}} + 1)^d/1$ became larger, which reaches the goal shown in Eq. (4).

To sum up, sampling the medoid objects as representative objects was sufficient to attain the goal shown in Eq. (4).

### *E. Neighborhood representative plus (NR+)*

Considering the case where a dataset contained global outlier only, for example a dataset with G1 and G3 in Fig. 2. (1), after using NR, the object in G1 was represented by an object in G3 as shown in Fig. 2. (4). Therefore, it shared the same outlier score with its representative object in G3, leading to the failure of the detectors in correctly classifying the object in G1 as an outlier. To solve this problem, we propose NR+, which is a revised version of NR. NR+ works in three steps: First, given an object $x_j$, it employs detector D to calculate the outlier score for $x_j$; Secondly it employs NR with detector D to calculate the outlier score for $x_j$; Thirdly, it sum up these two scores as the final outlier score for $x_j$. NR+ essentially attempts to balance the results between the object-level factor and group-level factor.

An example of using NR and NR+ was shown in Fig. 2. (5). The Fig. 2. (3) was the selected representative (medoid object) for the original data in Fig. 2. (1). The Fig. 2. (4) showed how each object was represented by medoid object nearby. The Fig. 2. (5) showed that G3 and G4 in the row labeled as *original* had similar outlier scores and G1, G2 and G3 in the row labeled as *NR* have similar outlier scores. However, there were no similar outlier scores in the row labeled as *NR+*. This improvement in result was derived from considering both object-level and group-level factors.

Since the process of finding representative could be repeated, NR and NR+ had two models when considering more repetitions. These two models could be illustrated by Fig. 9. The difference between these models lay in the decision to score the representative objects in each iteration versus the last iteration.

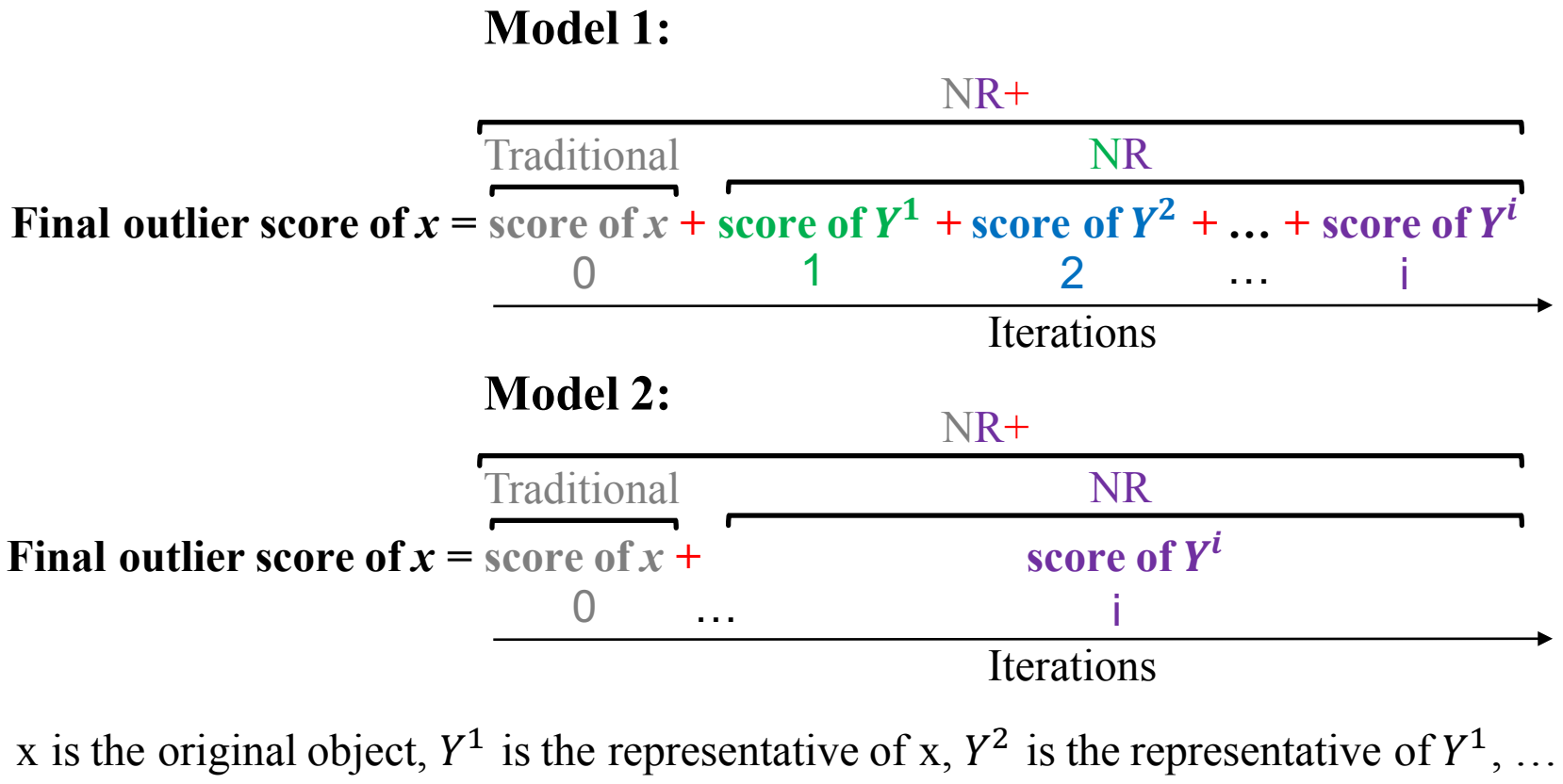

Fig. 9. Two models of considering *iteration.*

### *F. Discussion*

In terms of data representation, outlier detection and clustering share similarities. Outlier detection aims to find objects which can be represented by objects in other groups, while the clustering aims to find one representative object from each group of objects, such as *k*-medoid clustering algorithm. With their subtle but significant different, outlier detection could not be relegated to a subset of clustering [22].

According to the assumptions in Section III A, an object had a higher likelihood of being an outlier if its representative collective objects arose from another group of objects in data set. A group of objects had a higher chance of being classified as an outlier if they had the least representative objects in the dataset. Whether a group of objects were outliers or merely a cluster of normality objects depended on the definition and assumption in specific applications. However, the principle that the group having less objects were to be collective outliers still remained true, in view of the group-level factors.

The process to select representative (medoid) objects was similar to the process of *k*-NN-guided sampling. Hence, both processes resembled the concept of subsampling [44]. However, NR was fundamentally different from subsampling mainly in three ways: First, subsampling is one of the ensemble methods, which is a post-processing technique. Subsampling attempts to combine multiple detection results from detectors trained on different subsets of data sampled, while NR is not a post-processing technique but a preprocessing technique. Secondly, the purpose of subsampling is to capitalize on the diversity of data to improve detection results [44], while NR aims to consider the group-level factor to improve the detection accuracy in terms of collective outliers. Thirdly, subsampling is only suitable for *k*-NN based outlier detectors, while NR is suitable for all score-based detectors.

The objective of considering the group-level factor was to detect collective outliers more effectively. However, if data had no collective outliers, it was not necessary to compute separate outlier scores for objects in different groups. In a situation without any collective outliers, NR or NR+ does not provide additional benefit, but does not compromise accuracy of results as long as the threshold value of outlier scores were adjusted. In an unknown dataset, NR and NR+ thus provided greater accuracy without possibility of compromising outcome.

## IV. Experimental setup

Logically, a larger number of outliers in any dataset leads to a higher probability of presence of collective outliers. To evaluate the effectiveness of the proposed algorithms NR and NR+, eleven public real-world semantically meaningful datasets were selected [30]. The datasets were pre-processed by removing duplicates, subtracting the mean, and dividing by the standard deviation.

The performance of the outlier detectors was evaluated by the area under the *receiver operating characteristic* (ROC) curve. ROC values range from 0 to 1, where larger values equate better performance.

The selection of *k* value and evaluation of detectors were referenced to the article by Campos et al. [30]. The proposed neighborhood representative (NR) was tested with all values of *k* from 3 to 100. All *k*-NN-based outlier detectors were tested by setting the *k* to be equal to the *k* of NR. The best result of varying *k* was chosen for comparison for each *k*-NN-based outlier detector. For the other detectors, we used their default parameter settings found in their respective literature.

Tale I AUC FOR DATASETS WITH FEW OUTLIER

| | Detectors | **Shuttle** 1.30% | **WBC** 2.20% | **WDBC** 2.70% | **Lym.** 4.10% | **Glass** 4.20% | AVG | DIFF |
|---|---|---|---|---|---|---|---|---|
| Original | MOD+ | 0.96 | 0.99 | 1.00 | 0.85 | 0.86 | 0.93 | - |
| | LOF | 0.96 | 0.99 | 0.99 | 0.82 | 0.87 | 0.92 | - |
| | ODIN | 0.94 | 0.97 | 0.98 | 0.89 | 0.73 | 0.90 | - |
| | NC | 0.88 | 0.89 | 0.96 | 0.83 | 0.78 | 0.87 | - |
| | KNN | 0.96 | 0.99 | 0.99 | 0.86 | 0.87 | 0.94 | - |
| | MCD | 0.96 | 0.98 | 0.93 | 0.79 | 0.78 | 0.89 | - |
| | IFOREST | 0.82 | 0.99 | 0.93 | 0.86 | 0.78 | 0.87 | - |
| | OCSVM | 0.34 | 0.98 | 0.68 | 0.82 | 0.44 | 0.65 | - |
| | PCAD | 0.94 | 0.98 | 0.92 | 0.82 | 0.69 | 0.87 | - |
| | SO_GAAL | 0.01 | 0.99 | 0.50 | 0.60 | 0.21 | 0.46 | - |
| | Beta-VAE | 0.94 | 0.98 | 0.87 | 0.88 | 0.76 | 0.89 | - |
| | COPOD | 0.82 | 0.99 | 0.97 | 0.83 | 0.76 | 0.87 | - |
| | AVG | 0.79 | 0.98 | 0.89 | 0.82 | 0.71 | 0.84 | - |
| NR | MOD+ | 0.82 | 0.98 | 1.00 | 0.81 | 0.86 | 0.89 | -0.04 |
| | LOF | 0.93 | 0.97 | 0.99 | 0.80 | 0.79 | 0.90 | -0.02 |
| | ODIN | 0.85 | 0.93 | 0.94 | 0.78 | 0.70 | 0.84 | -0.06 |
| | NC | 0.88 | 0.93 | 0.91 | 0.76 | 0.80 | 0.86 | -0.01 |
| | KNN | 0.91 | 0.99 | 0.99 | 0.80 | 0.84 | 0.91 | -0.03 |
| | MCD | 0.97 | 0.99 | 0.94 | 0.81 | 0.84 | 0.91 | 0.02 |
| | IFOREST | 0.93 | 0.99 | 0.96 | 0.86 | 0.88 | 0.92 | 0.05 |
| | OCSVM | 0.76 | 0.99 | 0.81 | 0.82 | 0.45 | 0.77 | 0.12 |
| | PCAD | 0.96 | 0.99 | 0.94 | 0.80 | 0.74 | 0.89 | 0.02 |
| | SO_GAAL | 0.63 | 0.98 | 0.95 | 0.68 | 0.82 | 0.81 | 0.35 |
| | Beta-VAE | 0.96 | 0.98 | 0.95 | 0.87 | 0.79 | 0.91 | 0.02 |
| | COPOD | 0.88 | 0.99 | 0.98 | 0.81 | 0.82 | 0.90 | 0.03 |
| | AVG | 0.87 | 0.98 | 0.95 | 0.80 | 0.78 | 0.87 | 0.03 |
| NR+ | MOD+ | 0.93 | 0.99 | 1.00 | 0.84 | 0.87 | 0.93 | >0.01 |
| | LOF | 0.96 | 0.99 | 1.00 | 0.83 | 0.87 | 0.93 | 0.01 |
| | ODIN | 0.93 | 0.98 | 0.97 | 0.86 | 0.75 | 0.90 | >0.01 |
| | NC | 0.84 | 0.95 | 0.97 | 0.83 | 0.82 | 0.88 | 0.01 |
| | KNN | 0.96 | 0.99 | 0.99 | 0.84 | 0.88 | 0.93 | -0.01 |
| | MCD | 0.98 | 0.98 | 0.97 | 0.83 | 0.82 | 0.92 | 0.03 |
| | IFOREST | 0.90 | 0.99 | 0.96 | 0.87 | 0.84 | 0.91 | 0.04 |
| | OCSVM | 0.75 | 0.99 | 0.85 | 0.82 | 0.46 | 0.77 | 0.12 |
| | PCAD | 0.96 | 0.99 | 0.96 | 0.87 | 0.74 | 0.90 | 0.03 |
| | SO_GAAL | 0.18 | 0.99 | 0.95 | 0.64 | 0.60 | 0.64 | 0.18 |
| | Beta-VAE | 0.96 | 0.99 | 0.93 | 0.93 | 0.77 | 0.92 | 0.03 |
| | COPOD | 0.86 | 0.99 | 0.99 | 0.82 | 0.80 | 0.89 | 0.02 |
| | AVG | 0.85 | 0.99 | 0.96 | 0.83 | 0.77 | 0.88 | 0.04 |

TABE II AUC FOR DATASETS WITH A LARGE NUMBER OF OUTLIERS

| | Detectors | **Stamps** 9.10% | **Cardio.** 22.20% | **Pima** 34.90% | **Spam.** 39.40% | **Heart.** 44.40% | **Park.** 75.40% | AVG | DIFF |
|---|---|---|---|---|---|---|---|---|---|
| Original | MOD+ | 0.94 | 0.52 | 0.76 | 0.59 | 0.77 | 0.68 | 0.71 | - |
| | LOF | 0.89 | 0.59 | 0.69 | 0.49 | 0.67 | 0.60 | 0.66 | - |
| | ODIN | 0.84 | 0.61 | 0.63 | 0.52 | 0.61 | 0.53 | 0.62 | - |
| | NC | 0.68 | 0.57 | 0.57 | 0.55 | 0.58 | 0.61 | 0.59 | - |
| | KNN | 0.91 | 0.55 | 0.73 | 0.57 | 0.68 | 0.66 | 0.68 | - |
| | MCD | 0.85 | 0.49 | 0.68 | 0.41 | 0.64 | 0.65 | 0.62 | - |
| | IFOREST | 0.88 | 0.68 | 0.68 | 0.61 | 0.66 | 0.51 | 0.67 | - |
| | OCSVM | 0.87 | 0.70 | 0.62 | 0.53 | 0.58 | 0.43 | 0.62 | - |
| | PCAD | 0.90 | 0.75 | 0.63 | 0.55 | 0.62 | 0.38 | 0.64 | - |
| | SO_GAAL | 0.16 | 0.56 | 0.30 | 0.41 | 0.22 | 0.24 | 0.32 | - |
| | Beta-VAE | 0.87 | 0.76 | 0.55 | 0.55 | 0.38 | 0.23 | 0.55 | - |
| | COPOD | 0.93 | 0.66 | 0.65 | 0.69 | 0.69 | 0.54 | 0.70 | - |
| | AVG | 0.81 | 0.62 | 0.63 | 0.54 | 0.59 | 0.50 | 0.62 | - |
| NR | MOD+ | 0.90 | 0.57 | 0.70 | 0.65 | 0.71 | 0.86 | 0.73 | 0.02 |
| | LOF | 0.89 | 0.58 | 0.65 | 0.55 | 0.64 | 0.84 | 0.69 | 0.03 |
| | ODIN | 0.85 | 0.61 | 0.60 | 0.52 | 0.59 | 0.81 | 0.66 | 0.04 |
| | NC | 0.89 | 0.58 | 0.61 | 0.60 | 0.76 | 0.62 | 0.68 | 0.09 |
| | KNN | 0.93 | 0.57 | 0.72 | 0.64 | 0.77 | 0.87 | 0.75 | 0.07 |
| | MCD | 0.94 | 0.55 | 0.71 | 0.73 | 0.76 | 0.72 | 0.74 | 0.12 |
| | IFOREST | 0.95 | 0.73 | 0.70 | 0.70 | 0.72 | 0.59 | 0.73 | 0.06 |
| | OCSVM | 0.94 | 0.76 | 0.64 | 0.58 | 0.70 | 0.66 | 0.71 | 0.09 |
| | PCAD | 0.95 | 0.77 | 0.65 | 0.59 | 0.72 | 0.67 | 0.72 | 0.08 |
| | SO_GAAL | 0.89 | 0.76 | 0.42 | 0.43 | 0.35 | 0.70 | 0.59 | 0.27 |
| | Beta-VAE | 0.93 | 0.80 | 0.59 | 0.59 | 0.42 | 0.34 | 0.61 | 0.06 |
| | COPOD | 0.94 | 0.63 | 0.70 | 0.71 | 0.81 | 0.65 | 0.74 | 0.04 |
| | AVG | 0.92 | 0.66 | 0.64 | 0.61 | 0.66 | 0.69 | 0.70 | 0.08 |
| NR+ | MOD+ | 0.94 | 0.53 | 0.75 | 0.62 | 0.74 | 0.71 | 0.72 | 0.01 |
| | LOF | 0.90 | 0.59 | 0.69 | 0.50 | 0.66 | 0.64 | 0.67 | 0.01 |
| | ODIN | 0.90 | 0.64 | 0.64 | 0.53 | 0.62 | 0.55 | 0.64 | 0.02 |
| | NC | 0.90 | 0.58 | 0.61 | 0.58 | 0.75 | 0.59 | 0.67 | 0.08 |
| | KNN | 0.95 | 0.55 | 0.73 | 0.59 | 0.77 | 0.68 | 0.71 | 0.03 |
| | MCD | 0.94 | 0.52 | 0.72 | 0.70 | 0.78 | 0.68 | 0.72 | 0.10 |
| | IFOREST | 0.94 | 0.72 | 0.70 | 0.67 | 0.71 | 0.57 | 0.72 | 0.05 |
| | OCSVM | 0.93 | 0.74 | 0.65 | 0.56 | 0.62 | 0.49 | 0.67 | 0.05 |
| | PCAD | 0.95 | 0.77 | 0.66 | 0.59 | 0.72 | 0.67 | 0.73 | 0.09 |
| | SO_GAAL | 0.56 | 0.68 | 0.35 | 0.42 | 0.22 | 0.44 | 0.45 | 0.13 |
| | Beta-VAE | 0.93 | 0.79 | 0.58 | 0.58 | 0.37 | 0.25 | 0.58 | 0.03 |
| | COPOD | 0.94 | 0.65 | 0.69 | 0.72 | 0.79 | 0.59 | 0.73 | 0.03 |
| | AVG | 0.90 | 0.65 | 0.65 | 0.59 | 0.64 | 0.57 | 0.67 | 0.05 |

TABLE III AUC FOR GLASS WITH DIFFERENT *K* SELECTING METHODS

| | | NR | | | | | NR+ | | | | |
|---|---|---|---|---|---|---|---|---|---|---|---|
| Detectors | Original | Best | 3N | 5LOG | 6LOG | MNG | Best | 3N | 5LOG | 6LOG | MNG |
| MOD+ [36] | 0.86 | 0.86 | 0.87 | 0.87 | **0.88** | 0.86 | 0.87 | **0.88** | 0.84 | 0.87 | 0.86 |
| LOF [20] | 0.87 | 0.79 | 0.74 | **0.84** | 0.83 | 0.75 | 0.87 | 0.75 | **0.84** | 0.80 | 0.79 |
| ODIN [27] | 0.73 | 0.70 | **0.73** | 0.71 | 0.71 | 0.58 | 0.75 | **0.80** | 0.72 | 0.67 | 0.71 |
| NC [29] | 0.78 | 0.80 | 0.58 | 0.67 | 0.64 | **0.79** | 0.82 | 0.77 | 0.71 | 0.51 | **0.79** |
| KNN [28] | 0.87 | 0.84 | **0.87** | 0.86 | 0.86 | 0.59 | 0.88 | **0.89** | 0.83 | 0.86 | 0.61 |
| MCD [31] | 0.78 | 0.84 | **0.84** | 0.78 | 0.82 | 0.67 | 0.82 | **0.82** | 0.76 | 0.81 | 0.74 |
| IFOREST [32] | 0.78 | 0.88 | **0.75** | 0.72 | 0.72 | 0.72 | 0.84 | 0.78 | 0.77 | 0.78 | **0.81** |
| OCSVM [21] | 0.44 | 0.45 | 0.62 | 0.49 | 0.57 | **0.71** | 0.46 | **0.61** | 0.56 | 0.60 | 0.60 |
| PCAD [37] | 0.69 | 0.74 | **0.73** | 0.44 | 0.67 | 0.70 | 0.74 | **0.73** | 0.49 | 0.68 | 0.70 |
| SO_GAAL [40] | 0.21 | 0.82 | **0.79** | 0.42 | 0.53 | 0.78 | 0.60 | 0.60 | 0.71 | 0.81 | **0.89** |
| Beta-VAE [39] | 0.76 | 0.79 | **0.74** | 0.49 | 0.70 | 0.64 | 0.77 | 0.75 | 0.61 | 0.69 | **0.76** |
| COPOD [41] | 0.76 | 0.82 | **0.76** | 0.65 | 0.73 | 0.74 | 0.80 | 0.76 | 0.71 | 0.74 | **0.79** |
| AVG | 0.71 | 0.78 | 0.75 | 0.66 | 0.72 | 0.71 | 0.77 | 0.76 | 0.71 | 0.73 | 0.75 |
| Best | - | 4 | 8 | 1 | 1 | 2 | 2 | 6 | 1 | 0 | 5 |

TABLE IV AUC FOR GLASS WITH DIFFERENT *ITERATIONS*

| | | Model 1 | | | | Model 2 | | | |
|---|---|---|---|---|---|---|---|---|---|
| | | **NR** | | **NR+** | | **NR** | | **NR+** | |
| **Iteration** | **0** | **2** | **3** | **2** | **3** | **2** | **3** | **2** | **3** |
| MOD+ [36] | 0.87 | 0.83 | 0.71 | 0.88 | 0.86 | 0.83 | 0.79 | 0.89 | 0.87 |
| LOF [20] | 0.74 | 0.53 | 0.62 | 0.75 | 0.87 | 0.53 | 0.58 | 0.74 | 0.82 |
| ODIN [27] | 0.73 | 0.74 | 0.64 | 0.80 | 0.75 | 0.74 | 0.75 | 0.79 | 0.80 |
| NC [29] | 0.58 | 0.74 | 0.45 | 0.77 | 0.51 | 0.74 | 0.75 | 0.80 | 0.78 |
| KNN [28] | 0.87 | 0.82 | 0.74 | 0.89 | 0.87 | 0.82 | 0.78 | 0.89 | 0.87 |
| MCD [31] | 0.78 | 0.83 | 0.84 | 0.81 | 0.83 | 0.83 | 0.83 | 0.81 | 0.83 |
| IFOREST [32] | 0.78 | 0.81 | 0.84 | 0.84 | 0.86 | 0.83 | 0.81 | 0.83 | 0.83 |
| OCSVM [21] | 0.44 | 0.63 | 0.80 | 0.61 | 0.75 | 0.63 | 0.72 | 0.62 | 0.71 |
| PCAD [37] | 0.69 | 0.72 | 0.82 | 0.72 | 0.80 | 0.72 | 0.74 | 0.72 | 0.74 |
| SO_GAAL [40] | 0.21 | 0.67 | 0.71 | 0.60 | 0.60 | 0.52 | 0.56 | 0.58 | 0.64 |
| Beta-VAE [39] | 0.76 | 0.76 | 0.82 | 0.75 | 0.79 | 0.76 | 0.79 | 0.76 | 0.78 |
| COPOD [41] | 0.76 | 0.80 | 0.75 | 0.79 | 0.76 | 0.80 | 0.78 | 0.80 | 0.78 |
| AVG | 0.68 | 0.74 | 0.73 | 0.77 | 0.77 | 0.73 | 0.74 | 0.77 | 0.79 |
| Best | 0 | 0 | 5 | 2 | 1 | 0 | 0 | 2 | 1 |

## V. RESULTS

### *A. The overall result*

According to the percentage of outliers, the eleven datasets were categorized into two groups. One group contained a large number of outliers (>9%) and another had few outliers. The AUC results were summarized in Table I and Table II. As shown by the cells with the highlighted background, for datasets with few outliers, NR and NR+ improved the *original* results from 0.84 to 0.87 and 0.88 on average, respectively. For datasets with large number of outliers, NR and NR+ improved the *original* results from 0.62 to 0.70 and 0.67 on average, respectively.

The column labeled *DIFF* illustrated the difference between the column labeled *AVG* of NR and NR+ and the original. NR+ has the potential to improve any single detector regardless of quantity of outliers. The only exception is the KNN detector for datasets with few outliers. NR could improve all detectors for datasets with large number of outliers. For datasets with few outliers, NR could improve all non-k-NN-based detector, but could not improve k-NN-based detectors. The limitations of NR and NR+ might indicate that NR has limited usage for k-NN based detectors for datasets with few outliers, while NR+ is more robust than NR. Hence, NR+ is recommended to be used.

An interesting phenomenon in Table II was that MOD+ with 0.71 AUC on average was the best detector originally without NR/NR+. However, upon application of the proposed framework, the best detectors were KNN with 0.75 AUC on average when NR was applied and PCAD and COPOD with 0.73 AUC on average when NR+ was applied. This indicated that with NR/NR+, poorly performed detectors could become competitive when jointly used with NR/NR+.

However, NR/NR+ did not work well on some datasets and base detectors as show in Table I, which might be a result of two factors decided by the dataset and base detector. The dataset factor was that different datasets might have different ratio of collective outliers, hence the weightage of the group-level factor and the object-level factor should be adjusted according to applications, not equally as assumed by the current NR+. The base detector factor was that the parameter settings for the original datasets and the datasets pre-processed by NR/NR+ should be tuned, instead of using the same parameter settings as the current NR/NR+.

### *B. Effect of k in NR*

Determining the optimal value of *k* for *k*-NN based method was a challenge and related discussion could be found in a survey by Yang et al. [45]. Gu et al. suggested to use *k*=[0.03*N] [46] (3N), Li et al. suggested *k* = 2*[5*log(N)/2] +1 [29] (5LOG), and Li et al. suggested *k* = 6*log(N) in [47] (6LOG) where the brackets refer to rounding operation and N refers to the dataset size. In contrast, Ning et al. [48] suggested to select *k* based on a mutual neighbor graph (MNG). These four methods were studied for the *Glass* datasets, summarized in Table III. The column labeled *Original* corresponds to results from original detectors without NR or NR+. The column labeled *Best* corresponded to results from detectors without NR or NR+ with the best AUC which varied *k* from 3 to 100. We observed that the 3N method by Gu et al. [46] provided the best results for both NR and NR+ for both average AUC and number of best AUC per dataset. We therefore recommend using the 3N method as the default method of selecting *k*. 

### C. Effect of iteration of NR

NR could be performed with more iterations. We experimented with varying the number of iterations from 0 to 3 to study the effect of repetition with setting the *k*=[0.03*N] [46]. The value of *iteration* = 0 corresponded to the original detector without NR or NR+. There were two models when considering the *iterations*, shown in Fig. 9. The AUC results of each detector evaluated for the *Glass* datasets with iteration of 0, 2, and 3 were summarized in Table IV. The results of iteration of 1 were shown in Table III and when *iteration* = 1, where the results are identical for model 1 and model 2. We concluded that results were similar regardless of model used, and that more iterations did not bring additional benefit. We therefore suggest to use *iteration* =1 and model 2 as default iteration and model. However, we observed that the optimal results for each detector can be obtained by tuning the mode and iteration.

### D. Computational complexity

NR was based on *k*-NN, which required O(N*logN) calculations using the KD-tree or the Ball-tree. Hence, finding medoid would require O($k^2$NlogN). Bagaria et al. [49] presented a method to find the medoid with O(N*logN). Here, we used KD-tree for low dimensional datasets (D<20) and Ball-tree for high dimensional datasets (D>=20). The algorithm was implemented using Python 3.7. It was experimented on a laptop using CPU Intel Core i7, 16GB RAM, and clock frequency 2.3GHz. Table V shows the computing time.

TABLE V COMPUTING TIME (S)

| *k* \ N | 100 | 1000 | 10000 | 100000 |
|---|---|---|---|---|
| 3 | 0.06 | 0.67 | 6.32 | 66.54 |
| 9 | 0.06 | 0.66 | 6.32 | 67.56 |
| 27 | 0.06 | 0.63 | 6.62 | 68.64 |
| 81 | 0.07 | 0.75 | 7.68 | 75.55 |

### E. Limitations

The parameter *k* was essential to the proposed NR, but similar to other *k*-NN-based techniques, the choice of *k* was a challenging problem as it depended on the data distribution and detectors used. For *k*-NN-based outlier detectors, we recommended using the same *k* value used in the detectors for the *k* of NR. A more effective and practical method of selecting the parameter *k* would be promising.

As NR only computed scores for representative objects directly, this might result in false negative when true outlier objects were located in the marginal area between two clusters such as the G1 in Fig. 2. (1). Since the concept of NR was to compute lower outlier scores for objects in clusters with fewer objects, another limitation was that this might cause false positives when true normality objects were from clusters with smaller sizes such as G3 in Fig. 2. (1). Due to these limitations of NR, Fig. 2. (5) illustrated that NR computed the same mean outlier scores for group G1, G2, and G3, which was counterproductive as the groups were essentially indistinguishable. However, NR+ proved to be a more balanced solution between the binary options of using or not using NR, each with its own limitations. Fig. 2. (5) shows NR+ could overcome the problems faced by NR. Therefore, NR+ was recommended for applications requiring high accuracy, while NR was recommended for situations involving abundant collective outliers. The difference in performance due to different applications highlights the challenge faced by the field, and the relevance of both NR and NR+ in related work.

## VI. CONCLUSION

We proposed a novel pre-processing framework based on *k*-NN called neighborhood representative (NR). For the *k*-NN-based detectors, the *k* value in NR can be set as the same as *k* value in the detector. The technique can be used to improve all tested outlier detectors especially when the data has a large number of outliers. Experiments showed that it could significantly improve all tested outlier detectors (+8% AUC improvement on average related to the original). Lacking more effective methods to find optimal values for the parameter *k* of NR was the main limitation of this work, which could be future area of focus.